\pdfoutput=1

\documentclass[11pt]{article}

\usepackage[]{EMNLP2023}

\usepackage{times}
\usepackage{latexsym}
\usepackage{amsmath}
\usepackage[export]{adjustbox}

\usepackage[T1]{fontenc}

\usepackage[utf8]{inputenc}

\usepackage{microtype}

\usepackage{inconsolata}
\usepackage{multirow}
\usepackage{booktabs}

\title{NLIP\_Lab-IITH Multilingual MT System for WAT24 MT Shared Task}

\author{Maharaj Brahma \hspace{1em} Pramit Sahoo \hspace{1em} Maunendra Sankar Desarkar \\
       Natural Language and Information Processing Lab 
       (NLIP)\\ Indian Institute of Technology Hyderabad \\ Hyderabad, India\\  \texttt{\{cs23resch01004,ai23mtech14004\}@iith.ac.in}, \texttt{maunendra@cse.iith.ac.in}}
       
\begin{document}
\maketitle
\begin{abstract}
This paper describes NLIP Lab's multilingual machine translation system for the WAT24 shared task on multilingual Indic MT task for 22 scheduled languages belonging to 4 language families. We explore pre-training for Indic languages using alignment agreement objectives. We utilize bi-lingual dictionaries to substitute words from source sentences. Furthermore, we fine-tuned language direction-specific multilingual translation models using small and high-quality seed data. Our primary submission is a 243M parameters multilingual translation model covering 22 Indic languages. In the IN22-Gen benchmark, we achieved an average chrF++ score of 46.80 and 18.19 BLEU score for the En-Indic direction. In the Indic-En direction, we achieved an average chrF++ score of 56.34 and 30.82 BLEU score. In the In22-Conv benchmark, we achieved an average chrF++ score of 43.43 and BLEU score of 16.58 in the En-Indic direction, and in the Indic-En direction, we achieved an average of 52.44 and 29.77 for chrF++ and BLEU respectively. Our model\footnote{Our code and models are available at \url{https://github.com/maharajbrahma/WAT2024-MultiIndicMT}} is competitive with IndicTransv1 (474M parameter model). 
\end{abstract}

\section{Introduction}
Multilingual Neural Machine Translation (MNMT) has shown remarkable success in building translation systems for world languages in a single model \cite{johnson-etal-2017-googles}. These successes have led researchers to increase the model capacity catering to hundreds of world languages \cite{fan2020beyond}, \cite{nllb2022}. It also led to multilingual translation models for particular languages under particular geographical groups such as Indic \cite{ramesh2022samanantar, gala2023indictrans}, African \cite{nekoto-etal-2020-participatory}. Indic languages are interesting, with diverse languages belonging to various language families and written scripts.
\par
This paper describes our system submission for the WAT 24 MultiIndic22MT task \cite{dabre-etal-2024-multiindic22mt}, which includes 22 scheduled Indian languages belonging to 4 language families across 12 written scripts. We participated in the constrained translation task. We explore an alignment agreement-based pre-training objective. Specifically, we substitute words from source sentences for equivalent words in a random language. The pre-training data consists of a sentence pair from the original data and code-switched augmented sentences. Our primary submission is a fine-tuned transformer-based multilingual model with 243M parameters. Experimental results show that our system achieves an average chrF++ score of 46.80 for the En-Indic direction in the IN22-Gen benchmark. We achieved an average chrF++ score of 46.80 and 18.19 BLEU score for the En-Indic direction. In the Indic-En direction, we achieved an average chrF++ score of 56.34 and 30.82 BLEU score. Compared with the IndicTransv2 model for Indic-Indic translation, our system lags most minor for \texttt{pan\_Guru-snd\_Deva} with 0.3 chrF++ scores. Due to computational constraints, we train our model on a reduced corpus.
\section{Dataset}

\subsection{Pre-training data}
\begin{figure*}[!h]
    \centering
    \includegraphics[width=0.75\linewidth]{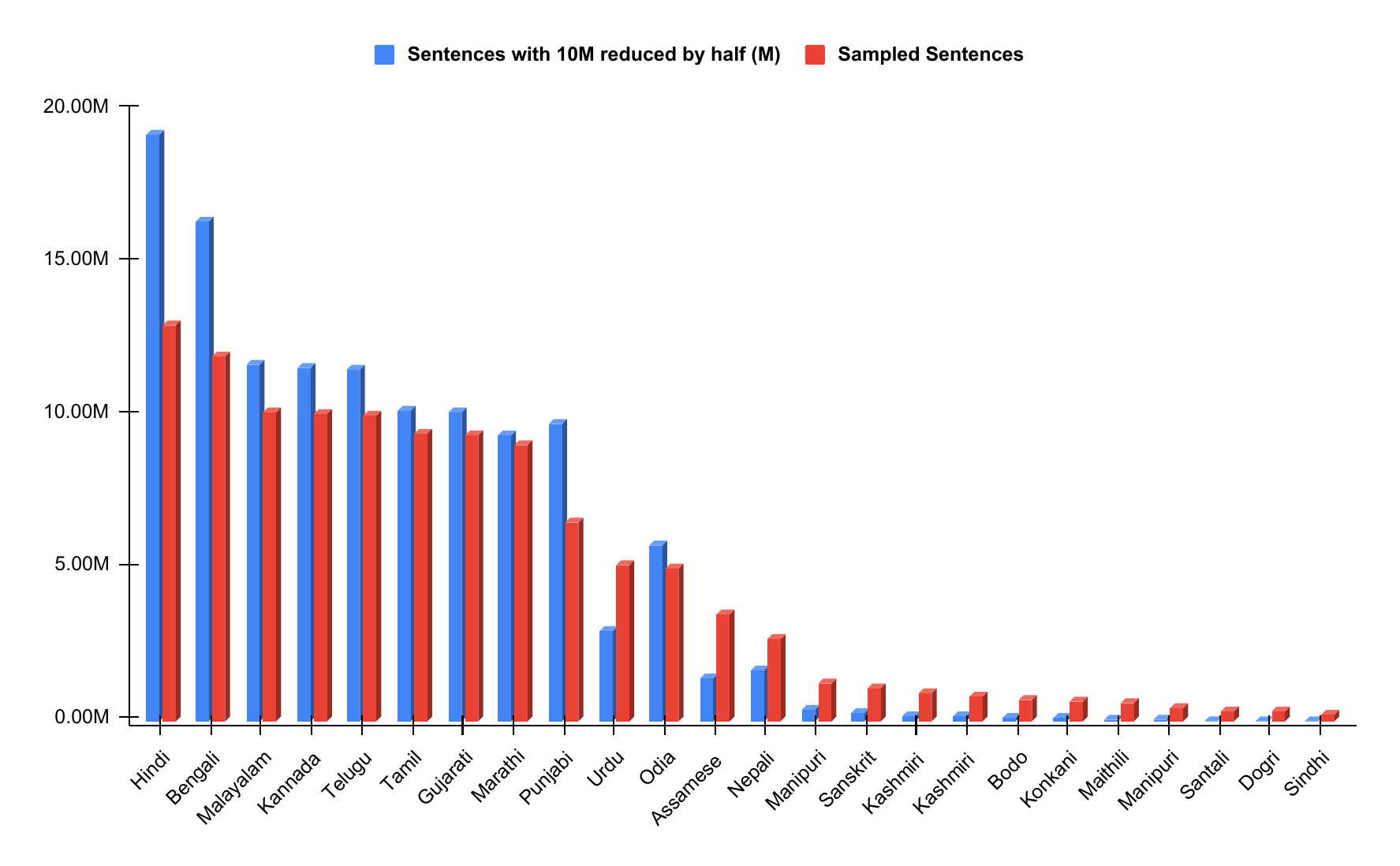}
    \caption{Number of sentences in each language and the sampled distribution with the T=5}
    \label{fig:sampledDistribution}
\vspace*{-0.2cm} 
\end{figure*}

In this section, we describe the dataset used for pre-training. We use the official Bharat Parallel Collection Corpus (BPCC) \cite{gala2023indictrans} but reduce the corpus size due to computational limitations. We also exclude sentences from the comparable directory. For languages with over 10 million parallel sentences, we reduce the no. of sentences by half. The corpus statistics are shown in Table \ref{tab:dataStatistics}. To handle skew data distribution and have good representation for low-resource languages, we use heuristic-based temperature sampling \cite{DBLP:journals/corr/abs-1907-05019, conneau-etal-2020-unsupervised} for data sampling with temperature sampling (T = 5) shown in Figure \ref{fig:sampledDistribution}. We utilize small, high-quality data from BPCC, namely ILCI, Massive, NLLB Seed, Daily, and Wiki, for direction-specific fine-tuning.

\begin{table*}[!h]
\centering
\scriptsize
\scalebox{0.97}{
\begin{tabular}{cccccc}
   \toprule
   \textbf{Language}  & \textbf{Script} & \textbf{\# of sentences (M)}  & \textbf{Language}  & \textbf{Script} & \textbf{\# of sentences (M)} \\
   \midrule
    Assamese & Bengali & 1.42 & Manipuri & Metei & 0.04 \\
    Bodo     & Devanagari  & 0.12 & Manipuri & Bengali & 0.37 \\
    Bengali  & Bengali     & 16.39 & Marathi &  Devanagari & 9.37 \\
    Dogri & Devanagari & 0.02 & Nepali & Devanagari & 1.68 \\ 
    Konkani & Devanagari & 0.10 & Odia & Oriya & 5.80 \\
    Gujarati & Gujarati & 10.12 & Punjabi & Gurmuki & 9.75 \\ 
    Hindi & Devanagari & 19.24 & Sanskrit & Devanagari & 0.28 \\ 
    Kannada & Kannada & 11.60 & Santali & Olck & 0.02 \\
    Kashmiri & Devanagari & 0.20 & Sindhi & Devanagari & 0.01 \\ 
    Kashmiri & Arabic & 0.15 & Tamil & Tamil & 10.18 \\
    Maithili & Devanagari & 0.09 & Telugu & Telugu & 11.54 \\
    Malayalam & Malayalam & 11.69 & Urdu & Arabic & 2.99 \\
    \bottomrule
\end{tabular}
}
\vspace*{-0.2cm} 
\caption{\small Statistics of the dataset. Total of 113.65 million bi-texts.}
\label{tab:dataStatistics}
\end{table*}

\subsection{Alignment Augmentation}
For alignment augmentation we English-centric bi-lingual dictionaries from MUSE\footnote{\url{https://github.com/facebookresearch/MUSE\#ground-truth-bilingual-dictionaries}} and GATITOS\footnote{\url{https://github.com/google-research/url-nlp/tree/main/gatitos}}. We use top 4000 words in dictionaries, replaced with a probability of 30\% from the bi-lingual dictionary. We consider only replacing words in the languages that have dictionaries.

\section{Methodology}
Our pre-training approach is inspired by aligning embeddings \cite{lin-etal-2020-pre, yang-etal-2020-csp} through substituting words from a bi-lingual dictionary. We pre-trained a universal model that covers En-Indic and Indic-En. We named this model \textbf{``IndicRASP''}. IndicRASP is fine-tuned into a language direction-specific model called \textbf{``IndicRASP-Seed''} using small and high-quality seed data.

\noindent\textbf{IndicRASP (IR)}: IndicRASP is pre-trained on data from 22 Indic languages sourced from BPCC. During pre-training, we randomly substitute English words for corresponding Indic language words, resulting in code-switched augmented sentences. The alignment augmentation technique helps to bring semantically similar embeddings closer together. 
We get 56M sentences after alignment augmentation. We combined training sentences from the original En-Indic and Indic-En\footnote{Reverse sentence pairs of En-Indic corpus} and obtained 282M sentences for pre-training.




\par
\noindent\textbf{IndicRASP-Seed (IR Seed)}: To further enhance the performance of IndicRASP, we fine-tuned the model to be language-direction specific. We consider high-quality seed data from the BPCC corpus: ILCI, NLLB Seed, Massive, Daily, and Wiki. We sampled a total of 2.26M sentences and fine-tuned IndicRASP for both En-Indic and Indic-En directions.

\section{Experiments}

\noindent\textbf{Setting:} We use the standard sequence-to-sequence Transformer big model as our architecture for pre-training. It uses 6 encoder and 6 decoder layers, with an embedding size of 1024. The embeddings between the encoder and decoder are shared, with a feed-forward network size of 4096 and 16 attention heads. 

\noindent\textbf{Training:} We pre-train the model with the Adam optimizer \cite{Kingma2014AdamAM} with $\beta_1$ set to 0.9 and $\beta_2$ set to 0.98. We set the warmup initial learning rate to 1e-07 and the learning rate to 5e-4, with a warmup step of 4000. We train the models with a dropout rate of 0.1 and a label smoothing rate of 0.1. During fine-tuning, we consider a learning rate of 3e-5 and a dropout rate of 0.2. All experiments are conducted on 8 NVIDIA A100 GPUs.

\noindent\textbf{Baseline Models:} We consider two baselines:
\begin{enumerate}
    \item \textbf{IndicTransv1 \cite{ramesh2022samanantar}}: IndicTransv1 (IT1) is a multilingual transformer \cite{vaswani2017attention} translation model for 11 Indic languages trained. It is a 474M parameter trained on the 49.7M sentence pair on the Samanantar dataset.
    \item \textbf{IndicTransv2 \cite{gala2023indictrans}}: IndicTransv2 (IT2) is a 1B parameter model trained on the BPCC corpus for 22 Scheduled Indian languages.
\end{enumerate} 

\noindent\textbf{Language-Direction Specific Models:} For our primary submission, we fine-tune IndicRASP with direction-specific small seed data for En-Indic and Indic-En. For the Indic-Indic model, we fine-tune the IndicRASP-Seed (En-Indic direction) on the Indic-Indic corpus extracted from the BPCC corpus.   

\noindent\textbf{Evaluation:} We use the dev set of BPCC IN-Gen as our validation and evaluate our model on the test set of BPCC IN-Gen and IN-Conv. We report our results on lexical-based automatic metrics BLEU \cite{papineni-etal-2002-bleu}, and chrF++ \cite{popovic-2017-chrf}. We use the sacreBLEU library for evaluation, with a chrF word order of 2. 


\begin{table}[!h]
\centering
\scriptsize
\scalebox{0.9}{
\begin{tabular}{l| ccc}
     \toprule
    \textbf{Language pair} & \textbf{BLEU}  & \textbf{chrF} & \textbf{chrF++}  \\
    \midrule
    asm\_Beng-eng\_Latn & 19.9 & 50.4 & 47.8 \\
    ben\_Beng-eng\_Latn & 22.1 & 50.1 & 48.0 \\
    brx\_Deva-eng\_Latn & 17.8 & 47.6 & 45.3 \\
    guj\_Gujr-eng\_Latn & 16.8 & 45.4 & 43.2 \\
    hin\_Deva-eng\_Latn & 23.1 & 50.5 & 48.5 \\
    kas\_Arab-eng\_Latn & 12.4 & 38.5 & 36.5 \\
    mal\_Mlym-eng\_Latn & 20.3 & 48.3 & 46.2 \\
    npi\_Deva-eng\_Latn & 18.0 & 46.7 & 44.5 \\
    san\_Deva-eng\_Latn & 9.3  & 34.7 & 32.6 \\
    sat\_Olck-eng\_Latn & 11.0 & 36.3 & 34.0 \\
    snd\_Deva-eng\_Latn & 21.2 & 47.1 & 45.5 \\
    tel\_Telu-eng\_Latn & 13.8 & 40.6 & 38.4 \\
    urd\_Arab-eng\_Latn & 20.3 & 45.6 & 43.9 \\
    \bottomrule
\end{tabular}}
\vspace*{-0.2cm} 
\caption{\small Indic-En scores results on hidden test set}
\label{tab:resultsHiddenTestSet}
\end{table}

\begin{table*}[!h]
\centering
\scriptsize
\begin{minipage}{0.45\textwidth}
\centering
\scalebox{0.72}{
    \begin{tabular}{l| cccc | cccc}
        \toprule
        \multirow{2}{*}{\textbf{Language}} 
            & \multicolumn{4}{c|}{\textbf{ En $-$ Indic}} & \multicolumn{4}{c}{\textbf{Indic $-$ En}}   \\ 
            & \textbf{IT1}  & \textbf{IT2}  & \textbf{IR} & \textbf{IR Seed} & \textbf{IT1}  & \textbf{IT2}  & \textbf{IR} & \textbf{IR Seed} \\
            \midrule
        asm\_Beng & 35.9 & 47.1 & 43.0 & 44.8 & 56.1 & 66.5 & 57.2 & 58.4 \\
        ben\_Beng & 48.6 & 51.8 & 47.3 & 48.5 & 58.4 & 64.5 & 55.6 & 56.9 \\
        brx\_Deva & --   & 47.8 & 46.3 & 46.5 & --   & 61.8 & 53.5 & 54.5 \\
        doi\_Deva & --   & 57.9 & 58.1 & 58.0 & --   & 72.7 & 64.0 & 64.5 \\
        gom\_Deva & --   & 45.2 & 42.1 & 42.9 & --   & 58.7 & 50.7 & 51.6 \\
        guj\_Gujr & 47.2 & 53.4 & 47.6 & 48.8 & 60.3 & 66.9 & 57.7 & 59.2 \\
        hin\_Deva & 53.3 & 56.6 & 52.4 & 54.8 & 60.7 & 65.0 & 57.6 & 58.7 \\
        kan\_Knda & 46.7 & 50.9 & 46.3 & 47.9 & 58.8 & 65.1 & 55.4 & 56.5 \\
        kas\_Arab & --   & 40.2 & 37.6 & 39.5 & --   & 60.5 & 52.6 & 53.8 \\
        mai\_Deva & --   & 48.7 & 46.7 & 47.3 & --   & 66.4 & 58.5 & 59.5 \\
        mal\_Mlym & 45.7 & 50.8 & 45.6 & 47.4 & 56.9 & 64.5 & 54.2 & 56.0 \\
        mni\_Mtei & --   & 44.5 & 44.5 & \textbf{45.0} & --   & 60.3 & 51.9 & 52.7 \\
        mar\_Deva & 44.3 & 50.9 & 44.2 & 46.7 & 57.7 & 65.1 & 55.8 & 57.3 \\
        npi\_Deva & --   & 49.0 & 44.8 & 47.8 & --   & 69.4 & 60.6 & 62.1 \\
        ory\_Orya & 40.3 & 43.8 & 43.0 & \textbf{46.1} & 60.0 & 67.6 & 57.6 & 59.3 \\
        pan\_Guru & 48.0 & 50.7 & 48.3 & 47.9 & 57.2 & 63.0 & 54.5 & 56.1 \\
        san\_Deva & --   & 38.6 & 34.6 & 36.2 & --   & 56.0 & 45.9 & 46.9 \\
        sat\_Olck & --   & 33.4 & 39.6 & \textbf{39.9} & --   & 47.7 & 47.2 & 48.2 \\
        snd\_Deva & --   & 36.5 & 34.2 & 35.2 & --   & 57.0 & 51.3 & 52.6 \\
        tam\_Taml & 45.5 & 49.6 & 45.4 & 46.4 & 53.9 & 59.7 & 51.3 & 53.2 \\
        tel\_Telu & 46.5 & 52.5 & 47.2 & 48.8 & 57.7 & 64.9 & 55.7 & 56.8 \\
        urd\_Arab & --   & 68.0 & 63.1 & 62.4 & --   & 73.1 & 63.5 & 64.7 \\
        \midrule
        \textbf{Avg.} & 45.64 & 48.54 & 45.50 & 46.80 & 57.97 & 63.47 & 55.10 & 56.34 \\
    \bottomrule
    \end{tabular}}
    \caption{\small chrF++ ($\uparrow$) scores on IN22-Gen}
    \label{tab:resultschrF2++In22Gen}
\end{minipage}
\begin{minipage}{0.45\textwidth}
\centering
\scalebox{0.72}{
\begin{tabular}{l| cccc | cccc }
    \toprule
    \multirow{2}{*}{\textbf{Language}} 
        & \multicolumn{4}{c|}{\textbf{ En $-$ Indic}} 
        & \multicolumn{4}{c}{\textbf{Indic $-$ En}}   \\ 
        & \textbf{IT1}  & \textbf{IT2}  & \textbf{IR} & \textbf{IR Seed} 
        & \textbf{IT1}  & \textbf{IT2}  & \textbf{IR} & \textbf{IR Seed} \\
        \midrule
    asm\_Beng & 9.9   & 19.3 & 15.0 & 17.8 & 32.5   & 42.5 & 30.7 & 31.6 \\
    ben\_Beng & 18.1   & 20.7 & 15.6 & 17.2 & 33.4   & 40.9 & 29.2 & 30.3 \\
    brx\_Deva & --   & 17.0 & 15.9 & 16.2 & --   & 39.0 & 27.2 & 28.2 \\
    doi\_Deva & --   & 33.8 & 33.7 & 33.4 & --   & 53.7 & 41.2 & 41.7 \\
    gom\_Deva & --   & 18.7 & 14.7 & 16.4 & --   & 34.0 & 23.8 & 24.9 \\
    guj\_Gujr & 17.9   & 25.6 & 18.2 & 19.6 & 36.3   & 43.5 & 31.3 & 32.6 \\
    hin\_Deva & 28.3   & 33.5 & 27.0 & 28.0 & 36.1   & 40.4 & 29.8 & 30.6 \\
    kan\_Knda & 13.4   & 17.7 & 13.0 & 15.6 & 34.8   & 40.5 & 29.0 & 30.0 \\
    kas\_Arab & --   & 14.4 & 12.4 & 13.4 & --   & 38.6 & 28.3 & 29.5 \\
    mai\_Deva & --   & 19.2 & 17.0 & 17.8 & --   & 43.2 & 32.8 & 33.8 \\
    mal\_Mlym & 13.9   & 16.4 & 12.0 & 13.1 & 31.4   & 41.0 & 28.2 & 30.1 \\
    mni\_Mtei & --   & 17.4 & 17.5 & 18.2 & --   & 39.0 & 27.7 & 28.9 \\
    mar\_Deva & 13.9   & 21.4 & 13.8 & 17.5 & 33.5   & 41.9 & 29.8 & 31.1 \\
    npi\_Deva & --   & 16.8 & 12.6 & 15.6 & --   & 48.2 & 35.7 & 38.0 \\
    ory\_Orya & 10.2   & 14.4 & 12.3 & 17.4 & --   & 45.1 & 31.4 & 32.6 \\
    pan\_Guru & 23.5   & 25.8 & 23.7 & 22.6 & 33.5   & 41.1 & 29.5 & 30.9 \\
    san\_Deva & --   & 10.9 & 8.4 & 9.1 & --   & 31.9 & 20.6 & 21.8 \\
    sat\_Olck & --   & 5.5 & 8.7 & 8.8 & --   & 25.1 & 23.1 & 24.3 \\
    snd\_Deva & --   & 13.9 & 10.1 & 11.1 & --   & 33.4 & 25.8 & 27.0 \\
    tam\_Taml & 11.9   & 14.7 & 11.3 & 11.7 & 28.9   & 36.1 & 25.6 & 27.1 \\
    tel\_Telu & 15.5   & 19.7 & 15.3 & 16.2 & 33.5   & 42.5 & 30.5 & 31.5 \\
    urd\_Arab & --   & 49.4 & 41.8 & 43.4 & --   & 53.8 & 40.1 & 41.6 \\
    \midrule
    \textbf{Avg.} & 16.0 & 20.28 & 16.82 & 18.19 & 30.0 & 40.7 & 29.60 & 30.82 \\
\bottomrule
\end{tabular}}
\caption{\small BLEU ($\uparrow$) scores on IN22-Gen}
\label{tab:resultsBLEUIn22Gen}
\end{minipage}
\end{table*}
\begin{table*}[!h]
    \centering
    \scriptsize
    \begin{minipage} {0.45\textwidth}
        \centering
        \scalebox{0.72}{
        \begin{tabular}{l | cccc | cccc}
            \toprule
            \multirow{2}{*}{\textbf{Language}} 
                & \multicolumn{4}{c|}{\textbf{ En $-$ Indic}} & \multicolumn{4}{c}{\textbf{Indic $-$ En}}   \\ 
                & \textbf{IT1}  & \textbf{IT2}  & \textbf{IR} & \textbf{IR Seed} & \textbf{IT1}  & \textbf{IT2}  & \textbf{IR} & \textbf{IR Seed} \\
                \midrule
            asm\_Beng & 36.4   & 46.8 & 40.9 & 44.9 & 52.5   & 62.9 & 52.7 & 57.7 \\
            ben\_Beng & 47.5   & 49.7 & 45.1 & 47.6 & 55.2   & 58.4 & 51.7 & 55.3 \\
            brx\_Deva & --   & 45.3 & 43.8 & 44.2 & --   & 56.3 & 50.1 & 50.9 \\
            doi\_Deva & --   & 53.9 & 55.4 & 55.2 & --   & 65.0 & 59.1 & 59.9 \\
            gom\_Deva & --   & 42.5 & 39.8 & 39.9 & --   & 51.7 & 46.6 & 47.3 \\
            guj\_Gujr & 49.1   & 53.1 & 46.9 & 48.5 & 56.9   & 62.0 & 54.7 & 58.1 \\
            hin\_Deva & 48.6   & 49.6 & 48.0 & 48.2 & 57.4   & 60.1 & 54.8 & 56.7 \\
            kan\_Knda & 32.6   & 33.8 & 31.7 & 32.3 & 44.0   & 47.5 & 40.4 & 43.9 \\
            kas\_Arab & --   & 35.6 & 28.7 & 34.3 & --   & 52.6 & 45.9 & 47.6 \\
            mai\_Deva & --   & 44.3 & 39.8 & 43.0 & --   & 57.8 & 52.3 & 52.9 \\
            mal\_Mlym & 43.8   & 45.7 & 41.7 & 42.9 & 50.6   & 54.3 & 47.2 & 50.7 \\
            mni\_Mtei & --   & 40.2 & 40.8 & 41.1 & --   & 52.5 & 48.5 & 49.1 \\
            mar\_Deva & 43.7   & 48.6 & 42.2 & 44.7 & 54.2   & 58.5 & 50.9 & 55.2 \\
            npi\_Deva & --   & 51.5 & 44.4 & 49.9 & --   & 63.0 & 56.0 & 59.1 \\
            ory\_Orya & 38.9   & 40.2 & 39.1 & 41.6 & 55.6   & 60.3 & 52.4 & 56.6 \\
            pan\_Guru & 54.0   & 57.8 & 53.1 & 54.1 & 58.1   & 62.7 & 54.8 & 58.5 \\
            san\_Deva & --   & 35.5 & 29.3 & 33.5 & --   & 48.3 & 40.2 & 42.6 \\
            sat\_Olck & --   & 34.6 & 41.7 & 41.7 & --   & 43.5 & 46.4 & 47.4 \\
            snd\_Deva & --   & 30.3 & 31.8 & 33.2 & --   & 49.6 & 49.5 & 50.1 \\
            tam\_Taml & 37.7   & 39.1 & 37.4 & 38.3 & 44.1   & 45.8 & 40.8 & 43.6 \\
            tel\_Telu & 42.5   & 45.5 & 40.8 & 42.4 & 48.5   & 52.9 & 45.8 & 49.3 \\
            urd\_Arab & --   & 61.6 & 54.6 & 53.9 & --   & 65.5 & 57.4 & 61.2 \\
            \midrule
            \textbf{Avg.} & 43.16 & 44.78 & 41.66  & 43.43  &  52.46 & 53.22 & 49.92 &  52.44  \\
        \bottomrule
        \end{tabular}}
        \caption{\small chrF++ ($\uparrow$) scores on IN22-Conv}
        \label{tab:resultschrF2++In22Conv}
        
    \end{minipage}
    \begin{minipage}{0.45\textwidth}
        \centering
        \scriptsize
        \scalebox{0.72}{
        \begin{tabular}{l| cccc | cccc }
           \toprule
            \multirow{2}{*}{\textbf{Language}} 
                & \multicolumn{4}{c|}{\textbf{ En $-$ Indic}} & \multicolumn{4}{c}{\textbf{Indic $-$ En}}   \\ 
                & \textbf{IT1}  & \textbf{IT2}  & \textbf{IR} & \textbf{IR Seed} & \textbf{IT1}  & \textbf{IT2}  & \textbf{IR} & \textbf{IR Seed} \\
                \midrule
                    asm\_Beng & 11.6   & 19.7 & 15.3 & 18.5 & 31.3   & 43.8 & 31.8 & 36.7 \\
                    ben\_Beng & 20.1   & 21.3 & 17.5 & 19.1 & 32.9   & 36.4 & 29.0 & 32.2 \\
                    brx\_Deva & --   & 15.4 & 13.6 & 14.7 & --   & 35.5 & 26.8 & 27.9 \\
                    doi\_Deva & --   & 32.4 & 34.1 & 34.4 & --   & 45.6 & 36.8 & 38.1 \\
                    gom\_Deva & --   & 14.2 & 11.3 & 11.2 & --   & 29.9 & 23.2 & 23.7 \\
                    guj\_Gujr & 23.2   & 27.2 & 20.9 & 22.3 & 34.7   & 41.1 & 32.0 & 35.4 \\
                    hin\_Deva & 28.4   & 30.1 & 27.4 & 27.5 & 35.5   & 39.3 & 32.5 & 34.0 \\
                    kan\_Knda & 6.1   & 6.7 & 5.1 & 5.8  & 21.1   & 24.9 & 17.8 & 19.8 \\
                    kas\_Arab & --   & 11.3 & 6.5 & 9.4  & --   & 31.8 & 23.1 & 25.2 \\
                    mai\_Deva & --   & 18.9 & 15.3 & 18.0 & --   & 36.6 & 28.7 & 29.3 \\
                    mal\_Mlym & 11.1   & 11.3 & 9.1 & 9.4  & 27.6   & 31.6 & 23.8 & 27.4 \\
                    mni\_Mtei & --   & 14.2 & 14.6 & 15.2 & --   & 31.9 & 26.1 & 26.9 \\
                    mar\_Deva & 15.5   & 19.4 & 14.7 & 16.2 & 32.2   & 36.7 & 28.5 & 32.6 \\
                    npi\_Deva & --   & 21.2 & 14.3 & 19.4 & --   & 42.4 & 33.5 & 36.9 \\
                    ory\_Orya & 11.3   & 12.3 & 11.7 & 13.9 & 33.6   & 38.8 & 30.4 & 34.1 \\
                    pan\_Guru & 32.0   & 35.7 & 30.8 & 31.5 & 36.8   & 43.0 & 33.2 & 37.0 \\
                    san\_Deva & --   & 6.3 & 3.9 & 5.5  & --   & 26.1 & 17.8 & 19.5 \\
                    sat\_Olck & --   & 6.6 & 10.9 & 10.6 & --   & 23.1 & 23.7 & 25.0 \\
                    snd\_Deva & --   & 7.4 & 8.3 & 9.2  & --   & 27.5 & 26.5 & 27.2 \\
                    tam\_Taml & 7.7   & 7.6 & 7.2 & 7.2  & 20.8   & 22.7 & 18.0 & 19.7 \\
                    tel\_Telu & 12   & 14.1 & 10.9 & 11.2 & 26.3   & 31.0 & 23.6 & 26.3 \\
                    urd\_Arab & --   & 43.7 & 33.5 & 34.6 & --   & 45.9 & 35.7 & 40.0 \\
                    \midrule
                    \textbf{Avg.} & 16.27 & 18.14 & 15.31 & 16.58 & 30.25 &  33.36 & 27.39  & 29.77  \\
                \bottomrule
                \end{tabular}
                }
                \caption{\small BLEU ($\uparrow$) scores on IN22-Conv}
                \label{tab:resultsBLEUIn22Conv}
    \end{minipage}
\end{table*}

\section{Results}
We list the results of our model on the IN22-Gen in Table \ref{tab:resultschrF2++In22Gen}, \ref{tab:resultsBLEUIn22Gen} for chrF++ and BLEU, respectively. Similarly, Table \ref{tab:resultschrF2++In22Conv} and \ref{tab:resultsBLEUIn22Conv} results for chrF++ and BLEU in IN22-Conv. Table \ref{tab:resultsHiddenTestSet} shows the performance of our primary submission on a hidden test set. Our findings described for IN22-Gen are:

\begin{itemize}
    \item IndicRASP achieves an average chrF++ score of 45.50, and IndicRASP-Seed achieves 46.80 with an improvement of (+1.30) for the En-Indic direction. Similarly, IndicRASP achieves an average BLEU score of 16.82 and 18.19 for the IndicRASP-Seed. It suggests that fine-tuning small, high-quality language directions improves the alignment augmented IndicRASP model. We can observe similar results for Indic-En.
    \item By comparing IT1 and IndicRASP-Seed, we find that IndicRASP-Seed has a chrF++ improvement of +1.30 for En-Indic; however, in the Indic-En direction, IndicRASP-Seed is lagging behind by 1.63.
    \item By comparing IT2 and IndicRASP-Seed, we find that IndicRASP-Seed lags behind by 1.74 chrF++ scores for En-Indic direction.  In the Indic-En direction, the IndicRASP-Seed lags behind significantly by a 7.13 chrF++ score from IT2.
    \item For En-Indic languages highlighted in bold in Table \ref{tab:resultschrF2++In22Gen}, namely Manipuri, Oriya, and Santali, IndicRASP-Seed performs better than IndicTransv2 with chrF++ score difference of 0.5, 2.3, and 6.5 respectively.
    \item We observe that our setup performs better in the En-Indic direction than in Indic-En. This is possibly due to the reduction of the dataset.
\end{itemize}

In Table \ref{tab:resultschrF++IndicIndic}, we show the performance of IndicRASP-Seed for Indic-Indic direction in the IN22-Gen and IN22-Conv datasets. We observe that the IT2 is better than the IndicRASP-Seed in all language pairs, particularly for \texttt{mal\_Mlym-hin\_Deva}, IndicRASP-Seed lags highest behind by a 5.6 chrF++ score, and \texttt{pan\_Guru-snd\_Deva} lags behind by a 0.3 chrF++ score.
\begin{table}[!ht]
\centering
\scriptsize
\scalebox{0.9}{
\begin{tabular}{l| cc}
     \toprule
    \textbf{Language pair} & \textbf{IT2}  & \textbf{IR Seed}  \\
    \midrule
    \multicolumn{3}{c}{\textbf{IN22-Gen}} \\
    \midrule
    ben\_Beng-hin\_Deva & 48.7 & 44.0 (-4.7) \\
    hin\_Deva-ben\_Beng & 45.7 & 41.3 (-4.4) \\
    hin\_Deva-mal\_Mlym & 44.4 & 39.2 (-5.2) \\
    mal\_Mlym-hin\_Deva & 48.0 & 42.4 (-5.6) \\
    pan\_Guru-snd\_Deva & 30.8 & 30.5 (-0.3) \\
    snd\_Deva-pan\_Guru & 41.1 & 37.5 (-3.6) \\
    tam\_Taml-tel\_Telu & 43.5 & 38.3 (-5.2) \\
    tel\_Telu-tam\_Taml & 45.4 & 41.5 (-3.9) \\
    \midrule
    \multicolumn{3}{c}{\textbf{IN22-Conv}} \\
    \midrule
    ben\_Beng-hin\_Deva & 44.3 & 40.8  (-3.5) \\
    hin\_Deva-ben\_Beng & 44.0 & 39.2  (-4.8) \\
    hin\_Deva-mal\_Mlym & 40.9 & 36.8  (-4.1) \\
    mal\_Mlym-hin\_Deva & 40.8 & 37.6  (-3.2) \\
    pan\_Guru-snd\_Deva & 29.4 & 28.5  (-0.9) \\
    snd\_Deva-pan\_Guru & 43.8 & 40.8  (-3.0) \\
    tam\_Taml-tel\_Telu & 37.4 & 32.5  (-4.9) \\
    tel\_Telu-tam\_Taml & 36.6 & 33.5  (-3.1) \\
    \bottomrule
\end{tabular}}
\vspace*{-0.2cm} 
\caption{\small Indic-Indic chrF++ ($\uparrow$) scores results on IN22-Gen and IN22-Conv dataset}
\label{tab:resultschrF++IndicIndic}
\end{table}

\section{Conclusion}
This paper presents our system for the WAT24 shared task on the MultiIndic22MT 2024 Shared Task. We focus on a universal model using pretraining Indic languages with alignment augmentation and further obtaining direction-specific models using finetuning on small and high-quality seed data. We submit a competitive 243M parameter model covering 22 Indic languages that achieves a comparable performance with a 474M parameter model covering 11 languages.

\section*{Limitations}
The present study particularly focuses on pre-training objectives on a parallel corpus. However, techniques such as utilizing monolingual corpus \cite{pan-etal-2021-contrastive} along with alignment objective remain unexplored. Also, large language models can be potentially leveraged to generate datasets for low-resource Indic languages. Further, we restricted the alignment augmentation of substitute words from source sentences (English words). However, words from target sentences can also be substituted can explored.


\section*{Acknowledgements}
We thank the reviewer for the valuable feedback. We are also grateful to the Department of Computer Science and Engineering, Indian Institute of Technology Hyderabad, for providing the necessary computing resources to conduct the experiments.

\bibliography{emnlp2023}
\bibliographystyle{acl_natbib}




\end{document}